\documentclass{article}

\usepackage{PRIMEarxiv}

\usepackage[utf8]{inputenc} 
\usepackage[T1]{fontenc}    
\usepackage{hyperref}       
\usepackage{url}            
\usepackage{booktabs}       
\usepackage{amsfonts}       
\usepackage{nicefrac}       
\usepackage{microtype}      
\usepackage{lipsum}
\usepackage{fancyhdr}       
\usepackage{graphicx}       
\usepackage{boldline,multirow}
\usepackage{makecell}
\usepackage{hhline}
\usepackage{soul}
\usepackage{amsmath}
\usepackage{amssymb}
\usepackage{algorithm}
\usepackage{algorithmic}

\makeatletter
\newcommand{\vast}{\bBigg@{4}}
\newcommand{\Vast}{\bBigg@{5}}
\makeatother

\DeclareMathOperator{\s.t.}{s.t.}
\DeclareMathAlphabet{\mathcal}{OMS}{cmsy}{m}{n}

\graphicspath{{media/}}     

\title{Scheduling Planting Time Through Developing an Optimization Model and Analysis of Time Series Growing Degree Units
}

\author{
  Javad Ansarifar, Faezeh Akhavizadegan, Lizhi Wang \\
  Department of Industrial and Manufacturing Systems Engineering\\
  Iowa State University\\
  Ames, IA 50011, USA\\
  \texttt{\{javad, faezeh, lzwang\}@iastate.edu} \\
}

\begin{document}
\maketitle

\begin{abstract}
Producing higher-quality crops within shortened breeding cycles ensures global food availability and security, but this improvement intensifies logistical and productivity challenges for seed industries in the year-round breeding process due to the storage limitations. In the 2021 Syngenta crop challenge in analytics, Syngenta raised the problem to design an optimization model for the planting time scheduling in the 2020 year-round breeding process so that there is a consistent harvest quantity each week. They released a dataset that contained 2569 seed populations with their planting windows, required growing degree units for harvesting, and their harvest quantities at two sites. To address this challenge, we developed a new framework that consists of a weather time series model and an optimization model to schedule the planting time. A deep recurrent neural network was designed to predict the weather into the future, and a Gaussian process model on top of the time-series model was developed to model the uncertainty of forecasted weather. The proposed optimization models also scheduled the seed population's planting time at the fewest number of weeks with a more consistent weekly harvest quantity. Using the proposed optimization models can decrease the required capacity by 69\% at site 0 and up to 51\% at site 1 compared to the original planting time.
\end{abstract}


\section{Introduction}
Global food availability and sustainability are two of the most fundamental challenges due to the growing population and running out of agricultural land required to produce food for people and livestock \cite{godfray2010food,foley2011solutions, akhavizadegan2021integration}. Additional challenges include increasingly variable growing conditions and climate change \cite{ansarifar2021machine, ansarifar2020performance, tilman2011global,smith2013delivering}. Data-driven strategies increase and improve the productivity and sustainability of agriculture by proposing appropriate and adaptive management practices (e.g. planting, irrigation, fertilizing, tilling, harvesting, and management), scheduling the activities in the agriculture field, and breeding plants with the highest-yielding genetics \cite{twine2004effects,tilman2011global, ansarifar2020performance, ansarifar2021interaction}. Although applying new methods and analytical approaches helps seed industries to produce higher-quality crops within shortened breeding cycles, ultimately ensuring required food for global food security, it comes with an unprecedented new set of challenges.

Recently, this improvement intensifies logistical and productivity issues for seed industries in the year-round breeding process due to the storage capacity limitations and erratic and inconsistent weekly harvest quantities. One of the most important decisions in management practices is scheduling planting time, which has significant implications in field crops' development and productivity, crop model applications, and in acquiring adaptation strategies for future climate change. Although implementing an optimal planting time reduces the negative impact on the environment and maximizes crop yield \cite{akhavizadegan2021time, araya2012risk, may2004early}, it hinders seed industries by increasing the storage requirement and logistic risk incurred during the seed production in the year-round breeding process \cite{ansarifar2019new, cid2014crop}.

Seed industries use analysis of a suite of management practices to identify the optimum schedule among the possibilities for management practices, including planting, irrigation, fertilizing, tilling, and harvesting. However, the complexities among management practices decision, resources, availability of seed, uncertain environment, and policies and procedures have led to the necessity of proposing a decision-making framework for management practices that consider logistic and storage limitation, seed production process, and environmental uncertainty. The 2021 Syngenta crop challenge in analytics was launched to find a critical decision in sustainable agriculture to optimally schedule the planting of seeds to ensure that when ears are harvested, facilities are not over capacity and that there is a consistent number of ears each week.

Estimation of crop planting time has received special attention among industry players, researchers, and academic actors. In-depth reviews of proposed approaches in crop planning problems have been published by Lowe and Preckel \cite{lowe2004decision} and Ahumada and Villalobos \cite{ahumada2009application}. Several methods have been proposed and reported in the literature for addressing crop planting problems because of the complex and nonlinear relationships between planting time and profitability of agricultural products and uncertainties of the environment. The majority of studies in the literature review have been conducted research on the scheduling of crop planting time in farmers’ fields during the summer growing season. In contrast to their work, we schedule planting time of different seed populations to produce seed for farmers in the year-round breeding at a seed industry level.

Three planting time scheduling methods have been proposed in the literature, including pre-defined and constant planting time, mathematical programming models, and statistical analysis methods \cite{waha2012climate}. In the first approach, based on long term observations, the constant and fixed planting time is derived as representing typical average planting time \cite{fodor2010agro, cammarano2013adapting, drewniak2013modeling, elliott2015global, sacks2010crop}. The second approach is the application of mathematical programming by adjusting the farming system with different management practices to optimize the planting time scheduling with limited resources. Mathematical programming methods include a linear programming model \cite{pal2010priority}, genetic algorithm for a weighted sum method \cite{wang2020joint}, simple heuristic allocation policy \cite{boyabatli2019crop}, heuristic selection algorithm using automatic fuzzy clustering \cite{gadallah2014fuzzy}, Bayesian optimization \cite{akhavizadegan2021time, akhavizadegan2021risk}, a strict mathematical framework using fuzzy set theory \cite{sadovski2019method}, calibrated crop model using a genetic algorithm \cite{waongo2014crop}, and integration of demand fuzzy time series modeling and linear programming methods \cite{indriyanti2019using,vitadiar2018production}. The planting scheduling algorithm was developed to optimize planting time based on the nearest distance to customers and the availability of greenhouses and open fields \cite{putri2019development}. Closest to our model, Li et al. \cite{li2016uncertain} developed a fuzzy-based linear multi-objective programming model under uncertainty for crop planting structure planning. The third approach is statistical analysis methods that determine the best planting time by measuring the yield and other objective function response to planting time. They include a segmented-linear regression model \cite{egli2009regional}, analysis of variance \cite{baum2019planting,shang2020detection}, rule-based methods \cite{dobor2016crop,moore2014modelling}, STICS model \cite{brisson2003overview}, DSSAT model \cite{jones2003dssat}, CERES-Maize model \cite{tofa2020modeling}, calibrated RZWQM2 model \cite{anapalli2016climate}, APSIM model \cite{akhavizadegan2021time, baum2020impacts,keating2003overview,holzworth2014apsim}, progression model and simulation analysis \cite{yang2017spatially}, nonlinear model \cite{kessler2020soybean}, and the greatest likelihood of planting based on cumulative heat units \cite{niu2009reliability}.

The majority of studies have attempted to shed light on the planting time scheduling in farmers' fields during the growing season using different sets of tools and methods. However, seed companies need to know the planting time of seeds beyond the conventional growing season (for year-round breeding) to keep up with genetic improvement in the production cycle. Based on this literature gap, we focus on optimally scheduling the planting time of different seed populations in the year-round breeding process so that there is a consistent number of harvested ears each week. First, to address weather information uncertainty during the year-round breeding process, the weather must be predicted based on time-series analysis of historical weather information. Then, we developed the optimization model to schedule the planting of seed population within a few harvest weeks with a more consistent weekly harvest quantity. To show our model's performance, the 2021 Syngenta crop challenge data was used for our computational results in different cases.

\section{Problem definition}

\begin{figure}[H]
	\includegraphics[height=0.9in]{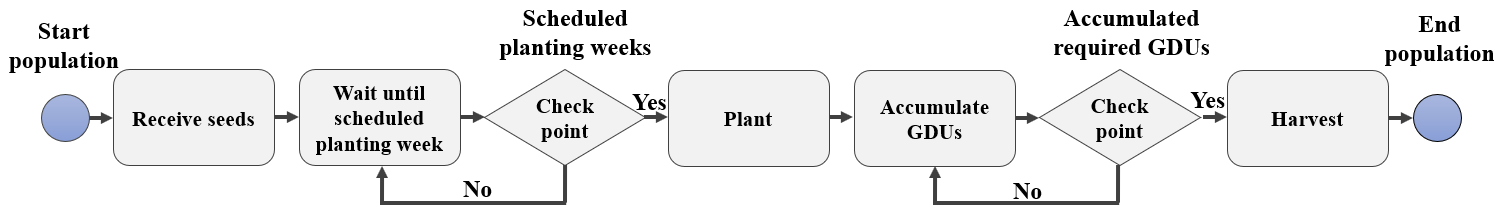}
	\caption{The year-round breeding process}\label{problem}
\end{figure}

The year-round breeding process of commercial corn as one of the world's most significant and planted crops is illustrated in Figure \ref{problem}. When seed populations arrived, they must wait for planting until their scheduled planting date is reached. After planting, they have to be mature enough to harvest. Growing degree units (GDUs) are a heuristic measurement in phenology that gardeners and farmers use to predict the crop development stages (e.g. emergence stage and maturity stage) by reaching the accumulated GDUs to a certain amount \cite{ miller2001using,wang1998simulation}. For several crops in different regions, a relationship between crop development stages and accumulated GDUs has been conducted \cite{chang2014simple, qian2009statistical,saiyed2009thermal}. GDUs are computed by taking the integral of warmth above a base temperature, approximately the average of the daily minimum and maximum temperatures.

\begin{figure}[H]
	\includegraphics[height=3in]{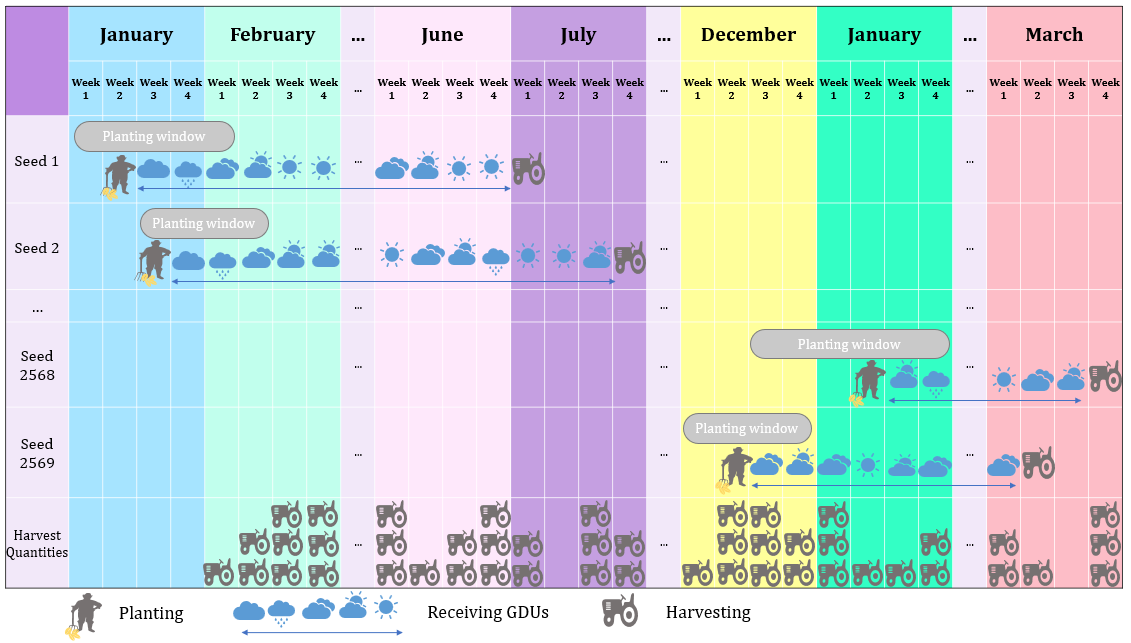}
	\caption{The year-round breeding process}\label{problem2}
\end{figure}

The goal of this paper is to develop the model to optimize the planting time of seed populations (see Figure \ref{problem2}) to address logistical and productivity challenges because of capacity limitations and inconsistent weekly harvest quantities.

\subsection*{Data}

The 2021 Syngenta crop challenge provided the Data, which includes seed populations, planting windows, required GDUs, harvest quantity, and historical GDUs information to optimize commercial corn's year-round breeding process.

\noindent \textbf{Calendar information:} This challenge is to schedule the seed populations' planting times during 2020. Week index is starting from the first week of January 2020. Each week runs from Sunday – Saturday.

\noindent \textbf{Seed population information:} This dataset includes 2569 seed populations, the planting site of each seed population, planting windows, the required number of GDUs in Celsius needed for the harvest, and the harvest quantity of each seed population. There are two different cases with specific population’s harvest quantity distribution. Syngenta simulated harvest quantities based on normal distributions for this challenge so that cases 1 and 2 follow normal distributions N(250,100) and N(350,150), respectively. However, in the real world, to estimate harvest quantities, we have to use predictive models based on historical information. Moreover, case 1 has the capacities, while there is no capacity limitation in case 2, and we are looking to determine the lowest possible capacity required. There are two different sites with different capacities. Site 0 has a capacity of 7,500 ears, and site 1 has a capacity of 6,000 ears at each week in case 1.

\noindent \textbf{Historical weather information:} This dataset includes historical GDUs in Celsius accumulated for each day for both sites during the last 11 years (2009-2019). Time series techniques can be used with this information to predict the GDUs in 2020. 

\subsection*{Objective function}

The objective of case 1 for the 2021 Syngenta crop challenge in analytics was to optimize each seed population's planting time at the fewest number of weeks so that the seed industry has a consistent weekly harvest quantity and capacity limitation constraint is met at each week. The objective of case 2 was to optimize each seed population's planting time so that the seed industry has a consistent weekly harvest quantity at the lowest possible capacity.

\section{Method}

\begin{figure}[H]
	\includegraphics[height=1.25in]{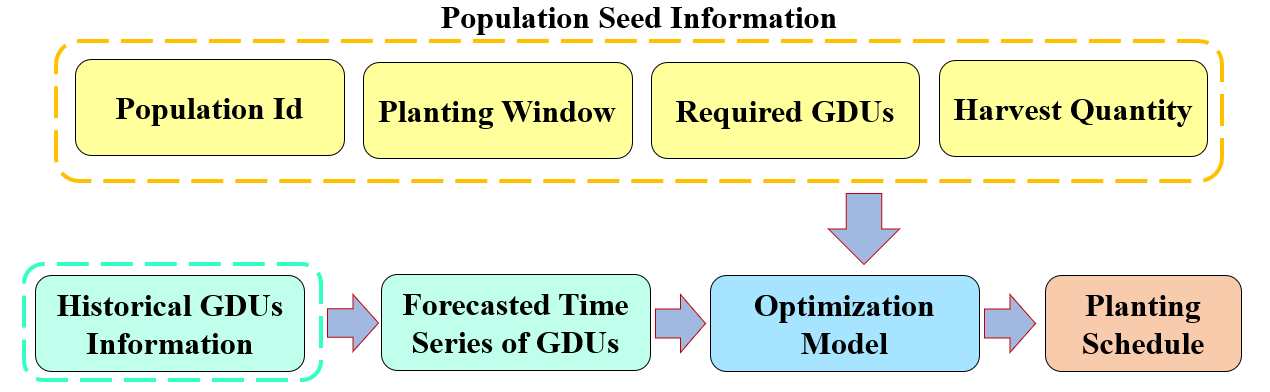}
	\caption{The overview of the proposed framework}\label{proposed}
\end{figure}

We developed a hybrid framework for this challenge, which combined the time series prediction model and optimization model to schedule the planting time of seed population in a one-year breeding process. The overview of this framework is diagrammed in Figure \ref{proposed}. This model includes two components: a weather prediction model that forecasts time series of GDUs for 2020 from historical GDUs information and an optimization model that finds optimal scheduling for planting seed populations to ensure a consistent weekly harvest. Details of two components are explained in the rest of this section.

\subsection*{Weather prediction model}

The harvesting date of seed population is determined based on accumulated GDUs in Celsius that seed population received during its planting time. Because the goal is to optimize the planting time for the next years (in this challenge is 2020 and 2021) and the weather information has not yet been observed, each site's GDUs for each calendar day of 2020 and 2021 must be predicted using the time series prediction model. Recent deep learning models have indicated high prediction accuracy in sequence processing and time series problems, particularly recurrent neural network. But, forecasting several steps into the future (in this problem for the next 2 years) is challenging with regards to keeping the forecasted weather within a reasonable range based on the historical information. Hence, we need a new predictive model to estimate uncertainty in the future. In this paper, a deep recurrent neural network (RNN) using long short-term memory(LSTM) network \cite{hochreiter1997long} was designed to capture nonlinear and temporal aspects of the GDUs. Moreover, to address the uncertainty of forecasted GDUs, we trained a Gaussian Process model \cite{rasmussen2003gaussian} to predict LSTM's residual errors. Details of LSTM's structure for the prediction of GDUs and the RIO model are described in the rest of this section.\\

\noindent\textbf{Prediction model design}

\begin{figure}[H]
	\includegraphics[height=1in]{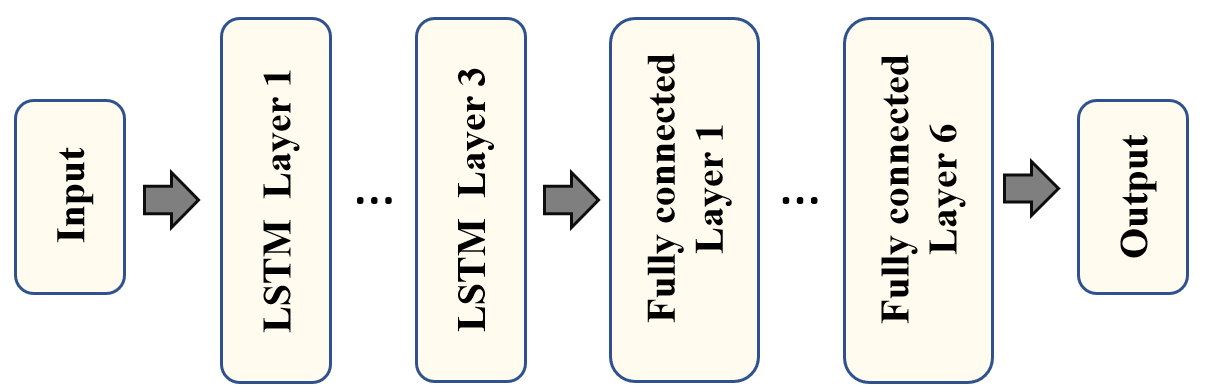}
	\caption{Outline of the predictive model structure.}\label{Structure}
\end{figure}

\begin{figure}[H]
	\includegraphics[height=1.5in]{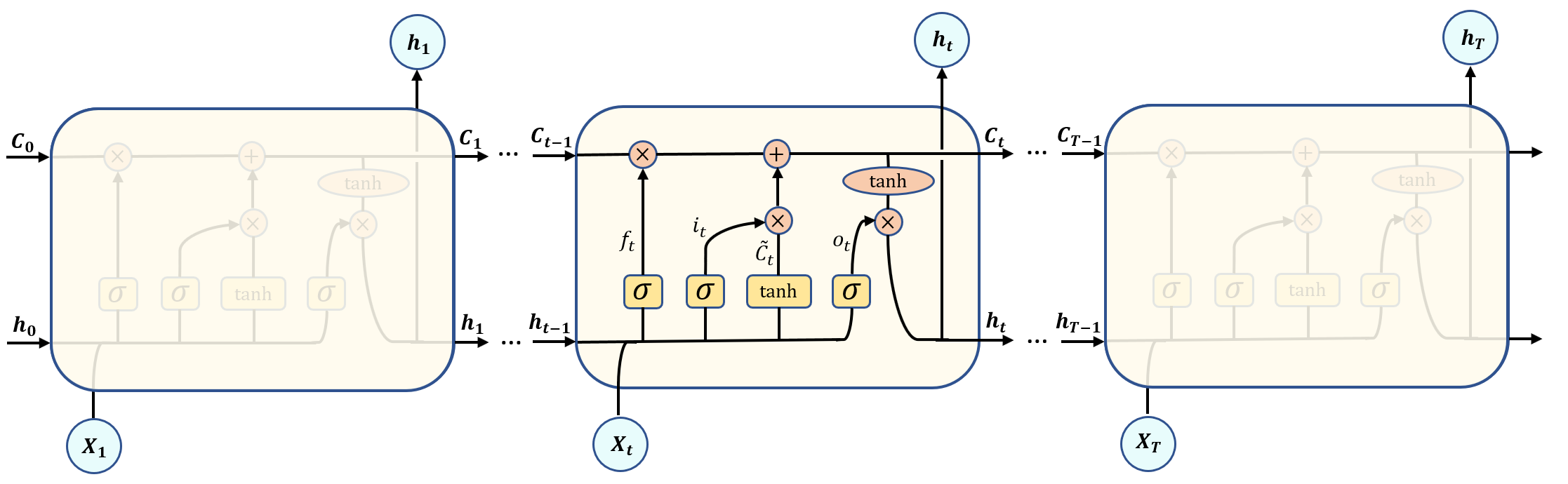}
	\caption{The overview of LSTM's structure. The first sigmoid layer is forget gate layer with output $f_t=\sigma(W_f.[h_{t-1},x_t]+b_f)$. The second sigmoid layer as part of the input gate layer has output $i_t=\sigma(W_i.[h_t-1,x_t]+b_i)$. The first tanh layer as part of the input generates a vector of new candidate values $\tilde{C}_t=tanh(W_c.[h_t-1,x_t]+b_c)$. The old cell state $C_{t-1}$ is calculated in the current cell $t$ by $C_t=f_t*C_{t-1}+i_t*\tilde{C}_t$. The third sigmoid layer as part of the output gate layer calculates output $o_t=\sigma(W_o.[h_{t-1},x_t]+b_o)$. The result of the output gate calculated by second tanh layer as $h_t=o_t*tanh(C_t)$.}\label{LSTM}
\end{figure}

We designed a new time-series prediction using LSTM network model and a fully connected neural network model to forecast GDUs using historical information. Figure \ref{Structure} indicates the outline of the proposed model. LSTM is an improved version of a RNN model employed widely to classify, process, and predict time-series problems. The main advantage of the LSTM over the conventional RNN model is that LSTM network solve the vanishing gradients problem because of using multiple gates instead of recurrent hidden neurons in their architectures. Also, the main difference between the LSTM models and conventional deep neural network is that LSTM is able to remember temporal dependency and patterns over time due to existing feedback connections in its structure. The structure of the LSTM is illustrated in Figure \ref{LSTM}.

In the LSTM structure, each time step has a cell with multiple gates as the cell's memory that manages, updates, and controls the flow of information throughout the network. The output of one cell at each time step is the next cell's input at the following time step. LSTM network contains three gates: forget gate, input gate, and output gate. The first sigmoid layer is known as the forget gate that is responsible for deciding what information must be kept and yield to cell state and what useless information must be forgotten. 
The input gate composes of the combination of the first tanh and the second sigmoid layers, which update the cell state with new encoded information. The output gate that consists of the second tanh and the third sigmoid layers controls the information flow. It decides and encodes part of the cell state as input in the following time step.

We used historical GDUs in Celsius accumulated for each day for both sites from 2009 to 2019 to train the time series prediction model by acquiring short term and long-term dependencies between sequence of GDUs. The previous 30 days of GDU are fed into the proposed time-series model to predict the GDU at the current day such that the proposed network is trained to predict GDU one day in the future. To make predictions far into the future, we can apply the trained model sequentially over the previous predicted GDU. A major limitation of this approach, however, is the dependency on previous predictions, which accumulates prediction errors over time  \cite{qiu2019quantifying}. In the next section, we introduce an auxillary model to improve the performance of the GDU prediction model by estimating and compensating the residual error. \\

\begin{algorithm}[H]
	\caption{\texttt{RIO Algorithm}} \label{RIO}
	\begin{algorithmic}[1]
		\STATE \textbf{Input:} Training set $D=\{(X,Y)\}=\{(x_i,y_i)\}_{i=1}^n$ where $x_i$ and $y_i$ is the previous 30 days of GDU at day $i$ and GDu at day $i$, $T$ as number of days for forecasting.  
		\STATE \textbf{Output:} $\tilde{Y}=\{\tilde{y}_i\}_{i=1}^T$ as the prediction of $T$ days far into the future.
		\STATE \textbf{Train phase}	
		\STATE Train the proposed network by feeding $D=\{(X,Y)\}$ and estimate $Y$ as $\hat{Y}$.
		\STATE Compute residual error $E=Y-\hat{Y}$.
		\STATE Feed $X$ back into the network and extract output of the last LSTM layer as $g(X)$.
		\STATE Train Gaussian process regression using new data set $\{g(X),E\}$ to estimate residual error for $x$ and its prediction $\hat{y}$ as Gaussian distribution  $\mathcal{N} (\bar{\hat{e}},\text{var}(\hat{e}))$, where
		\[ \bar{\hat{e}}=k((g(x),\hat{y}),(g(X),\hat{Y})) k((g(X),\hat{Y}),(g(X),\hat{Y}))^{-1} E \] \[\text{var}(\hat{e})=k((g(x),\hat{y}),(g(x),\hat{y}))-k((g(x),\hat{y}),(g(X),\hat{Y})) k((g(X),\hat{Y}),(g(X),\hat{Y}))^{-1} k((g(X),\hat{Y}),(g(x),\hat{y}))\]
		and $k$ denotes I/O kernel $k((x_i,\hat{y}_i),(x_i,\hat{y}_i))=\sigma_{\text{in}}^2\exp(-\frac{1}{2l^2_{\text{in}}}||x_i-x_j||^2)+\sigma_{\text{out}}^2\exp(-\frac{1}{2l^2_{\text{out}}}||\hat{y}_i-\hat{y}_j||^2)$ with the hyperparameters $\sigma_{\text{in}},l^2_{\text{in}},\sigma_{\text{out}},l^2_{\text{out}}$.
		\STATE \textbf{Forecast phase}
		\STATE Set $\hat{x}_1=Y[n-29:n]$.	
		\FOR{$t=1$ to $T$}		
		\STATE Feed $\hat{x}_t$ into the network and extract output of the last LSTM layer as $g(\hat{x}_t)$ and its prediction as $\hat{y}_t$.
		\STATE Use Trained Gaussian process regression model to compute $\bar{\hat{e}}$ and $\text{var}(\hat{e})$.
		\STATE The predicted GDU is sampled as follow $\tilde{y}_t \sim \mathcal{N}(\hat{y}_t+\bar{\hat{e}},\text{var}(\hat{e}))$.
		\STATE Set $\hat{x}_{t+1}=[\hat{x}_t[2:n],\tilde{y}_t]$. 
		\ENDFOR
	\end{algorithmic}
\end{algorithm}

\noindent\textbf{Modeling uncertainty in weather prediction}

The idea of the uncertainty estimation of the proposed model is to design another predictive model to estimate the residual error of the proposed network. To develop a more robust estimation far into the future, the Bayesian model can be integrated with the proposed time-series model to measure uncertainty \cite{neal2012bayesian, kim2019attentive, gal2016dropout, snoek2012practical}. This paper utilizes another machine learning model to predict the uncertainty directly by predicting residual error and augment the estimated error to the proposed model's prediction. This method is known as RIO (Residual estimation with an I/O kernel) \cite{qiu2019quantifying}. The RIO's structure is described in the algorithm \ref{RIO}. In this approach, a modified Gaussian Process regression model (GP) \cite{qiu2019quantifying} is trained to estimate the original residual errors in the training data set. This modified GP uses a new kernel (I/O kernel) that makes use of both inputs and outputs of the proposed time-series model to capture its behavior by estimating the residual error of the proposed time-series model. This I/O kernel is composed of the input kernel that corresponds to the training set and the output kernel that corresponds to the original model's prediction.

To construct the I/O kernel for our proposed model, the last LSTM layer's output and the predicted GDU from the proposed model are fed into the kernel of RIO as input kernel and output kernel. After training the modified GP with the I/O kernel, we estimate a Gaussian distribution for the residual error of the proposed time-series model such that we can compute both the mean and the standard deviation prediction of GDU. The future estimation of GDUs is calculated via Monte Carlo rollouts. Instead of predicting GDU at each day in the future and feeding the predicted value back into the proposed time-series model to predict the next day, we take a sample from the Gaussian distribution returned by RIO. Then this sample is fed back into the model to predict the next day. Sampling from the Gaussian distribution helps uncertainty estimation in the predictions, and we can create several weather scenarios by taking multiple samples from the Gaussian distribution. For this paper, we generated 25 weather scenarios by sampling 25 times from the Gaussian distribution of all days of two test years (2020 and 2021).

\subsection*{Optimization model}

Since the main objective is to optimize the planting time of seed population, we cast the scheduling problem as the optimization model using the predicted GDU as the heuristic measurement for harvesting. For case 1, the optimization model tried to schedule the planting of seed population at a minimum number of harvest weeks so that there is consistent harvest quantity among all weeks, and the capacity constraints are met. While at case 2, the optimization model determines the optimal scheduling of seed population's planting time at the lowest possible capacity required. Two sites do not have interaction with each other, and we can optimize them separately. Moreover, we developed one optimization problem for case 1 (when sites have storage capacity) and one optimization for case 2 (when sites do not have storage capacity). In the following, the variables and parameters used in the model are described.

\noindent \textbf{Sets and indices:}

\noindent\begin{tabular}{lp{5.9in}}
	$\mathcal{I}$ & Set of seed populations, $i\in \mathcal{I}=\{1,...,I\}$; \\
	$\mathcal{T}$ & Set of days in planting horizon, $t\in \mathcal{T}=\{1,...,T\}$; \\
	$\mathcal{W}$ & Set of weeks in planting horizon, $t\in \mathcal{W}=\{1,...,W\}$;\\
	$\mathcal{S}$ & Set of weather scenarios, $s\in \mathcal{S}=\{1,...,S\}$.\\
\end{tabular}\\

\noindent \textbf{Parameters:}

\noindent\begin{tabular}{lp{5.9in}}
	$\mathcal{R}_i$ & Accumulated growing degree units needed for harvesting seed population $i$;\\	
	$\mathcal{E}_i$ & Earliest date for planting seed population $i$;\\
	$\mathcal{L}_i$ & Latest date for planting seed population $i$;\\
	$\mathcal{H}_i$ & Harvest quantity (number of ears) for seed population $i$;\\
	$\mathcal{C}$ & Capacity of site for problem case 1;\\ 
	$\mathcal{P}_s$ & Probability of weather scenario $s$;\\
	$\mathcal{G}_{t,s}$ & GDUs during day $t$ at weather scenario $s$;
\end{tabular}\\

\noindent\begin{tabular}{lp{5.9in}}
	$\mathcal{M}_{w,t}$ & Binary parameter indicating whether the day $t$ belongs to week $w$ ($\mathcal{M}_{w,t}=1$) or not ($\mathcal{M}_{w,t}=0$);\\
	$y_{i,t,s,w}$ & Harvest quantity of seed population $i$ in week $w$ at weather scenario $s$ when it is planted in day $t$. $y_{i,t,s,w}$ is computed as $y_{i,t,s,w}=\mathcal{H}_i*\mathcal{M}_{w,t'} \forall i \in \mathcal{I}, t \in \mathcal{T}, \mathcal{E}_i\leq t\leq\mathcal{L}_i, s \in \mathcal{S}, w \in \mathcal{W}$ where $t'\leq T$ so that $\sum_{t''=t}^{t'-1} \mathcal{G}_{t'',s}\geq \mathcal{R}_i$ and $\sum_{t''=t}^{t'-1} \mathcal{G}_{t'',s}-\mathcal{R}_i\leq \mathcal{G}_{t,s}$, otherwise, $y_{i,t,s,w}=0$.
\end{tabular}\\

\noindent \textbf{Decision variables:}

\noindent\begin{tabular}{lp{5.9in}}
	$x_{i,t}$ & Binary variable indicating whether the seed population $i$ is planted in day $t$ ($x_{i,t}=1$) or not ($x_{i,t}=0$);\\
	$z$ & Auxiliary variable indicating maximum harvesting amount among all weeks and weather scenarios.
\end{tabular}\\

The mathematical programming model for case 1 is formulated as the following optimization model in Equations (\ref{obj1}-\ref{con3}).

\begin{align}
	\min& \sum_{s \in \mathcal{S}} \mathcal{P}_s \max_{w \in \mathcal{W}}\{\mathcal{C}-\sum_{i \in \mathcal{I}}\sum_{t \in \mathcal{T}} y_{i,t,s,w}x_{i,t}\} &\label{obj1}\\
	\s.t.&\sum_{t=\mathcal{E}_i}^{\mathcal{L}_i}x_{i,t} = 1 &\forall i \in \mathcal{I} \label{con1}\\
	&\sum_{i \in \mathcal{I}}\sum_{t \in \mathcal{T}}x_{i,t} = I & \label{con1p}\\
	&\sum_{i \in \mathcal{I}}\sum_{t \in \mathcal{T}} y_{i,t,s,w}x_{i,t} \leq \mathcal{C}   &\forall s \in \mathcal{S}, w \in \mathcal{W} \label{con2}\\
	&x_{i,t}=\{0,1\} & \forall i \in \mathcal{I}, t \in \mathcal{T} \label{con3}
\end{align}

The objective function in Equation (\ref{obj1}) is to minimize the expected maximum difference between the capacity and the weekly harvest quantity among all harvest weeks. Constraints (\ref{con1})  and (\ref{con1p}) specify the planting date of each seed population between their earliest and latest planting dates. Constraint (\ref{con2}) limits the weekly harvest quantity within existing capacity. Constraint (\ref{con3}) is the definition of binary decision variables.

After creating weather scenarios for the next planting and harvesting calendar, we solve the optimization model (\ref{obj1})-(\ref{con3}). It is better to use Equation (\ref{new}) instead of the Equation (\ref{obj1}) for objective function because it better reflects the fluctuation in weekly harvest quantity among all harvest weeks. Equation (\ref{new}) computes the difference of all pairs of the weekly harvest quantity among all harvest weeks. But using that makes the model intractable for the large size of the problem. Hence, we use the Equation (\ref{obj1}) as the objective function to solve the model within a reasonable time and make the model tractable for large problem. Since one of the evaluation criteria is to minimize the total number of harvest weeks, we iteratively shrink the available weeks for harvest so that the model (\ref{obj1})-(\ref{con3}) cannot result in harvesting in these weeks. Then, we calculate Equation (\ref{new}) just for the harvesting period (from the first harvest week to the last harvest week), and then we select the best period of harvesting time regarding minimizing Equation (\ref{new}). 

\begin{align}
	\sum_{s \in \mathcal{S}} \mathcal{P}_s\sum_{w \in \mathcal{W}}\sum_{w' \in \mathcal{W},w < w'}|\sum_{i \in \mathcal{I}}\sum_{t \in \mathcal{T}} y_{i,t,s,w}x_{i,t}-\sum_{i \in \mathcal{I}}\sum_{t \in \mathcal{T}} y_{i,t,s,w'}x_{i,t}| \label{new}
\end{align}

The optimization model for case 2 (model (\ref{obj2})-(\ref{con6})) is formulated as the same as case 1 by modifying objective function and the Constraint (\ref{con2}) to accommodate the model for the uncapacitated version. After finding the minimum capacity using model (\ref{obj2})-(\ref{con6}), the model (\ref{obj1})-(\ref{con3}) are applied to schedule the planting time of seed population. 

\begin{align}
	\min & \quad z &\label{obj2}\\
	\s.t.&\sum_{t=\mathcal{E}_i}^{\mathcal{L}_i}x_{i,t} = 1 &\forall i \in \mathcal{I} \label{con4}\\
	&\sum_{i \in \mathcal{I}}\sum_{t \in \mathcal{T}}x_{i,t} = I & \label{con4p}\\
	&\sum_{i \in \mathcal{I}}\sum_{t \in \mathcal{T}} y_{i,t,s,w}x_{i,t} \leq z   &\forall s \in \mathcal{S}, w \in \mathcal{W} \label{con5}\\
	&x_{i,t}=\{0,1\} & \forall i \in \mathcal{I}, t \in \mathcal{T} \label{con6}
\end{align}

The objective function in Equation (\ref{obj2}) is to minimize the maximum capacity required among all harvest weeks. Constraints (\ref{con4}) and (\ref{con4p}) assign the planting date of each seed population between their earliest and latest dates of planting. Constraint (\ref{con5}) calculates the maximum harvesting quantity among all harvest weeks. Constraint (\ref{con6}) is the definition of binary decision variables.

\subsection*{Experiment settings}

\begin{figure}[H]
	\includegraphics[height=2in]{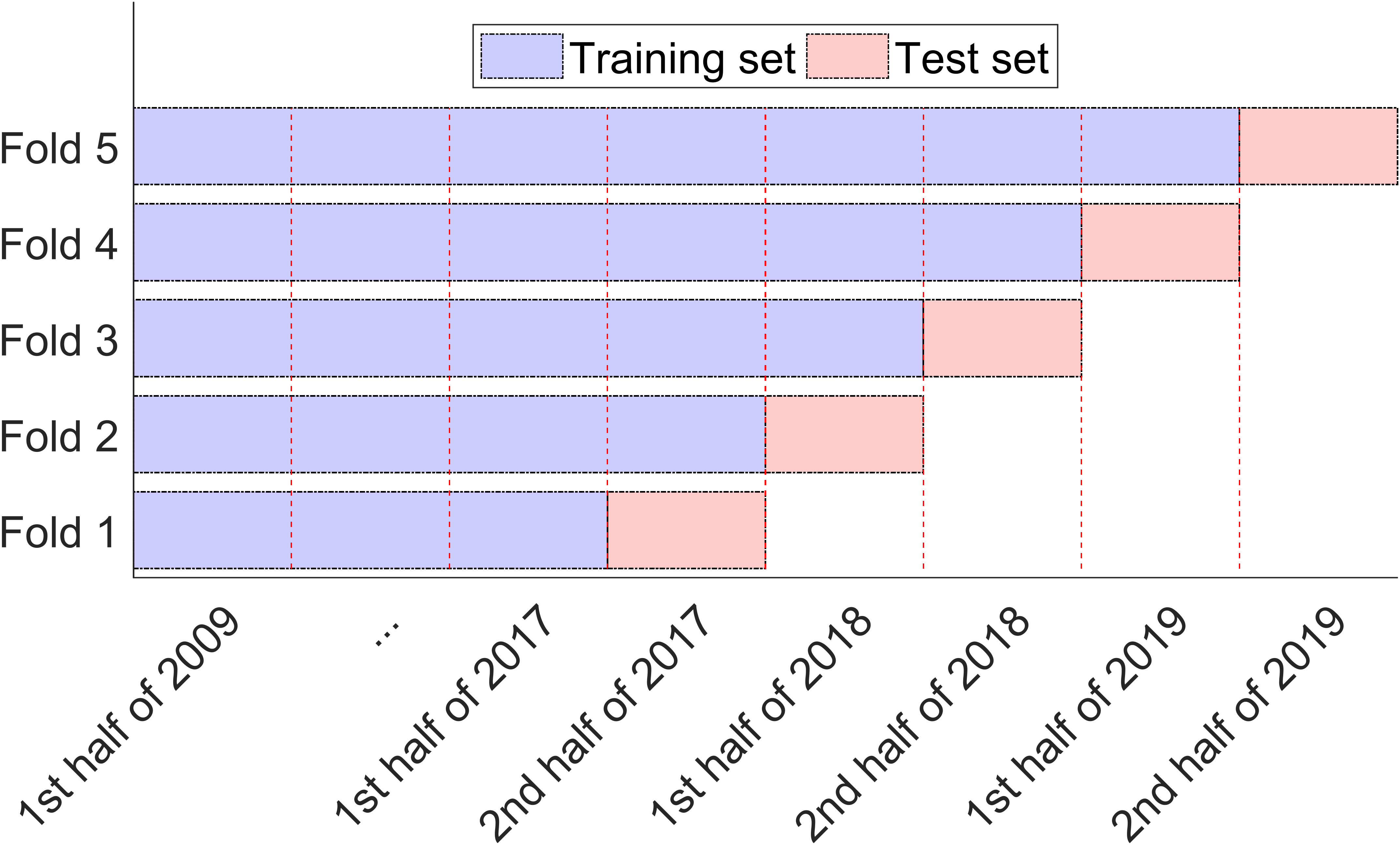}
	\caption{Partition of training and test data sets for cross-validation.}\label{CV}
\end{figure}

The proposed time-series model (both LSTM network and fully connected layers) was implemented in python using the TensorFlow package \cite{abadi2016tensorflow}. Parameter tuning of the hyperparameters of LSTM and fully connected layers indicated that the LSTM layer with 20 units and a dense layer with 20 neurons and a rectified linear unit (ReLU) activation function resulted in the most accurate model to capture the nonlinear and temporal aspects of the weather information. To tune the parameter, we used a time-wise five-fold cross-validation, as shown in Figure \ref{CV}. Each fold corresponds to particular six months as test data for prediction, and the rest of the data from 2009 up to test data corresponds to the training set. We applied Adam optimizer \cite{kingma2014adam} with a learning rate of 0.001 and a mini-batch size of 32. Adam optimizer tries to minimize mean absolute error (MAE) instead of mean squared error (MSE) because MAE is more robust in training the model with noisy training data.

To compare with the proposed time-series structure with the state-of-the-art, two different deep learning structures (convolutional neural network and deep fully connected neural network) were deployed. Details of CNN and DNN models are provided as follows.

\begin{itemize}
	\item \textbf{DNN}: DNN with 5 nonlinear layers is implemented in Python by using the TensorFlow package \cite{abadi2016tensorflow}. Each layer has 20 neurons and a ReLU activation function. We used the batch normalization \cite{ioffe2015batch} to increase the prediction accuracy. 
	\item \textbf{CNN}: CNN with three convolution layers and three max-pooling layers is implemented in Python by using the TensorFlow package \cite{abadi2016tensorflow}. The output of the last max-pooling layer is flattened and fed into two fully connected layers with 100 neurons and a ReLU activation function.
\end{itemize}

To tune the DNN parameters (numbers of hidden layers and neurons at each layer) and CNN parameters (numbers of convolution layers, fully connected layers and their neurons), we used a time-wise five-fold cross-validation (shown in Figure \ref{CV}) which led to the lowest cross-validation prediction error. Adam optimizer \cite{kingma2014adam} with a learning rate of 0.001, and a mini-batch size of 32 were applied to train the DNN and CNN model with regards to minimizing MAE.

The formulated optimization models were implemented in Python 3. Then, they were solved with the Gurobi optimizer version 9.0 \cite{optimization2014inc}.

\section{Quantitative results}

In this section, we report the computational experiments conducted in this research to test the proposed structure's performance in predicting weather into the future and the optimization models' performance in scheduling the planting time of seed population with more consistent harvest quantity among all weeks.

\subsection{Prediction accuracy comparison with other machine learning models}

We compared the performance of the proposed time-series structure with the state-of-the-art in terms of three criteria: RMSE, which indicates the difference between predicted and observed weather, relative RMSE (RRMSE), which represents the normalized difference between predicted and observed weather, and coefficient of determination ($R^2$), which computes the proportion of the variance in the weather that is explained by independent variables. Table \ref{tests} summarizes the daily benchmark of GDU prediction performance of the proposed structure and other models over five test years (2015-2019) to illustrate the impact of the proposed model. These results indicate that the proposed time-series model outperformed other machine learning models for all test years for both sites in all evaluation criteria.

\begin{table}[H]
	\centering
	\caption{Daily prediction performance of three time-series models for five test years (2015 to 2019) at sites 0 and 1.}\label{tests}
	\scalebox{0.9}{
		\begin{tabular}{cccccccc}
			\clineB{1-8}{2.5}
			\multirow{2}{*}{criterion}&\multirow{2}{*}{Site}&\multirow{2}{*}{Method}& &\multicolumn{3}{c}{Test Year}\\
			\hhline{~~~-----}
			&&&2015 &2016 &2017 &2018 &2019 \\
			\hline
			\multirow{6}{*}{RMSE}
			&\multirow{3}{*}{Site 0}&DNN&0.475&	0.493&	0.531&	0.509&	0.428\\
			&&CNN&0.585&	0.602&	0.632&	0.577&	0.544\\
			&&Proposed&0.429&	0.471&	0.476&	0.447&	0.404\\
			\hhline{~-------}
			&\multirow{3}{*}{Site 1}&DNN&0.718&	0.834&	0.882&	0.747&	0.702\\
			&&CNN&1.078&	1.106&	1.112&	1.054&	1.018\\
			&&Proposed&0.689&	0.727&	0.767&	0.710&	0.684\\
			\hline
			\multirow{6}{*}{RRMSE} 
			&\multirow{3}{*}{Site 0}&DNN&5.03\%	&5.33\%	&5.95\%	&5.84\%	&5.07\%\\
			&&CNN&6.21\%	&6.5\%	&7.08\%	&6.62\%	&6.43\%\\
			&&Proposed&4.55\%	&5.09\%	&5.33\%	&5.12\%	&4.77\%\\
			\hhline{~-------}
			&\multirow{3}{*}{Site 1}&DNN&6.74\%	&7.71\%	&8.42\%	&7.22\%	&6.5\%\\
			&&CNN&10.13\%&10.23\%	&10.61\%	&10.18\%	&9.42\%\\
			&&Proposed&6.47\%	&6.72\%	&7.32\%	&6.85\%	&6.33\%\\
			\hline
			\multirow{6}{*}{$R^2$}
			&\multirow{3}{*}{Site 0}&DNN&0.971	&0.972	&0.977	&0.971	&0.983\\
			&&CNN&0.956	&0.959	&0.968	&0.963	&0.973\\
			&&Proposed&0.976	&0.975	&0.982	&0.978	&0.985\\
			\hhline{~-------}
			&\multirow{3}{*}{Site 1}&DNN&0.744	&0.736	&0.709	&0.759	&0.714\\
			&&CNN&0.423	&0.536	&0.539	&0.522	&0.399\\
			&&Proposed&0.764	&0.799	&0.780	&0.783	&0.728\\
			\clineB{1-8}{2.5}
	\end{tabular}}
\end{table}

\begin{figure}[H]
	\includegraphics[height=2in]{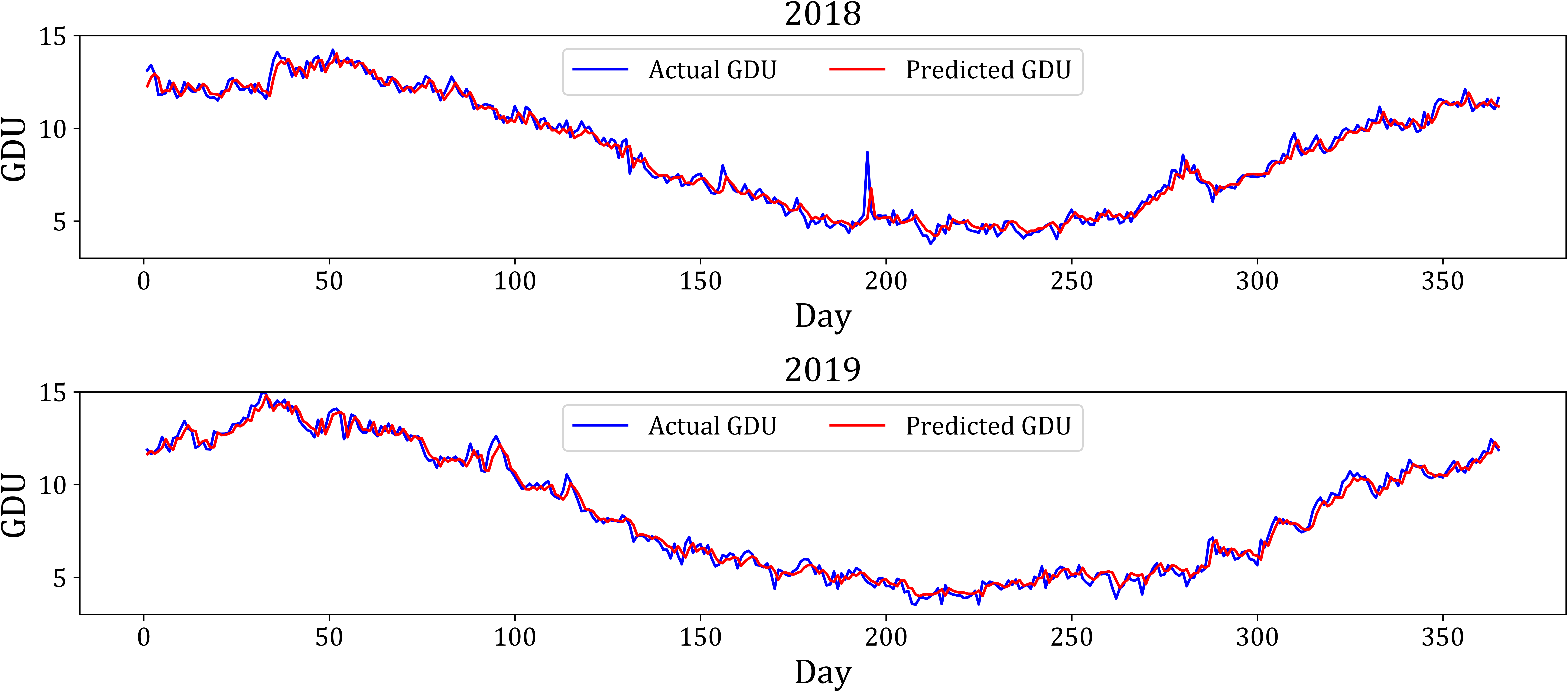}
	\caption{Daily GDU predictions of site 0 for test years 2018 and 2019.}\label{LSTMSITE0}
\end{figure}

\begin{figure}[H]
	\includegraphics[height=2in]{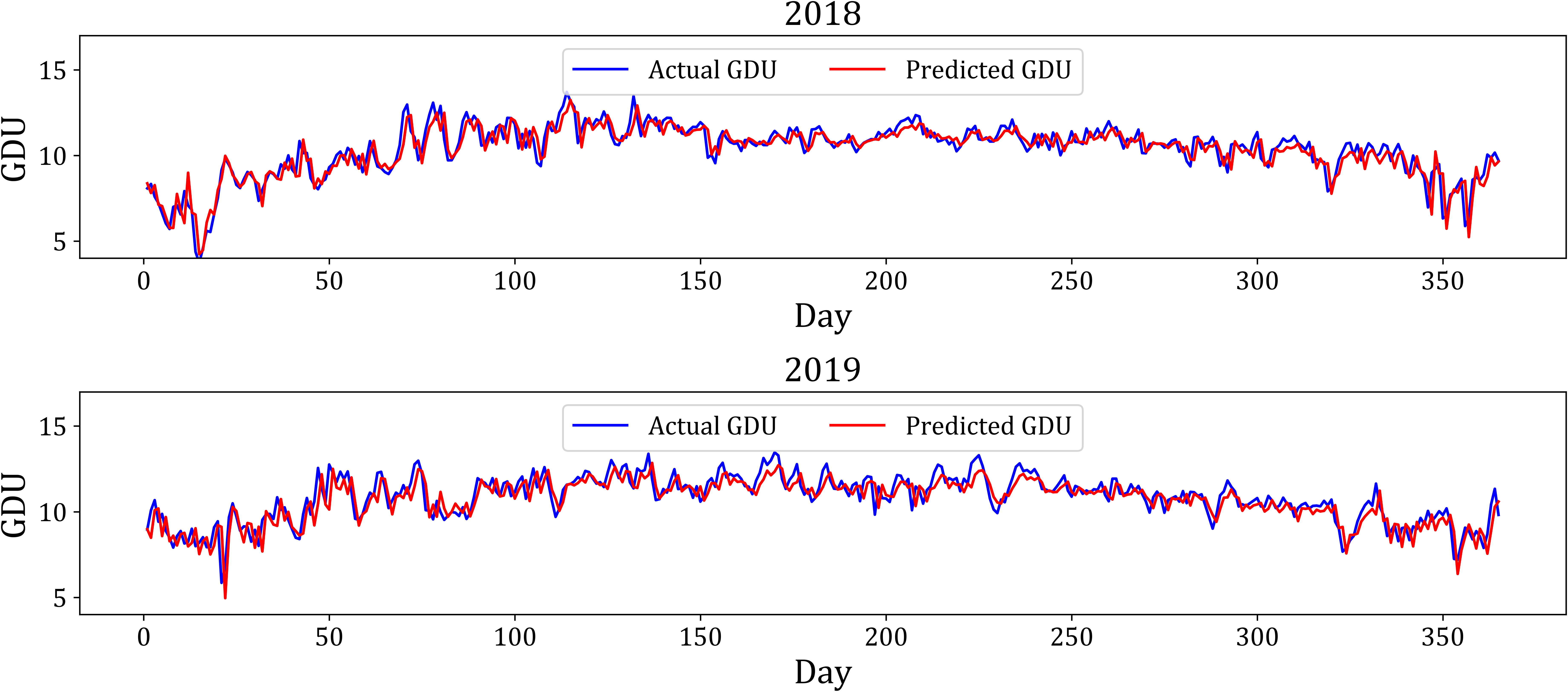}
	\caption{Daily GDU predictions of site 1 for test years 2018 and 2019.}\label{LSTMSITE1}
\end{figure}

Figures \ref{LSTMSITE0} and \ref{LSTMSITE1} illustrate the consistency of daily prediction of GDU with actual GDU at two test years (2018 and 2019) for sites 0 and 1, respectively. For this prediction, the previous 30 days of GDU are fed into the proposed time-series model to predict GDU one day in the future. The proposed model also shows its ability to capture both the overall trend of GDU over test years and GDU fluctuations from one day to another.

\begin{figure}[H]
	\includegraphics[height=2in]{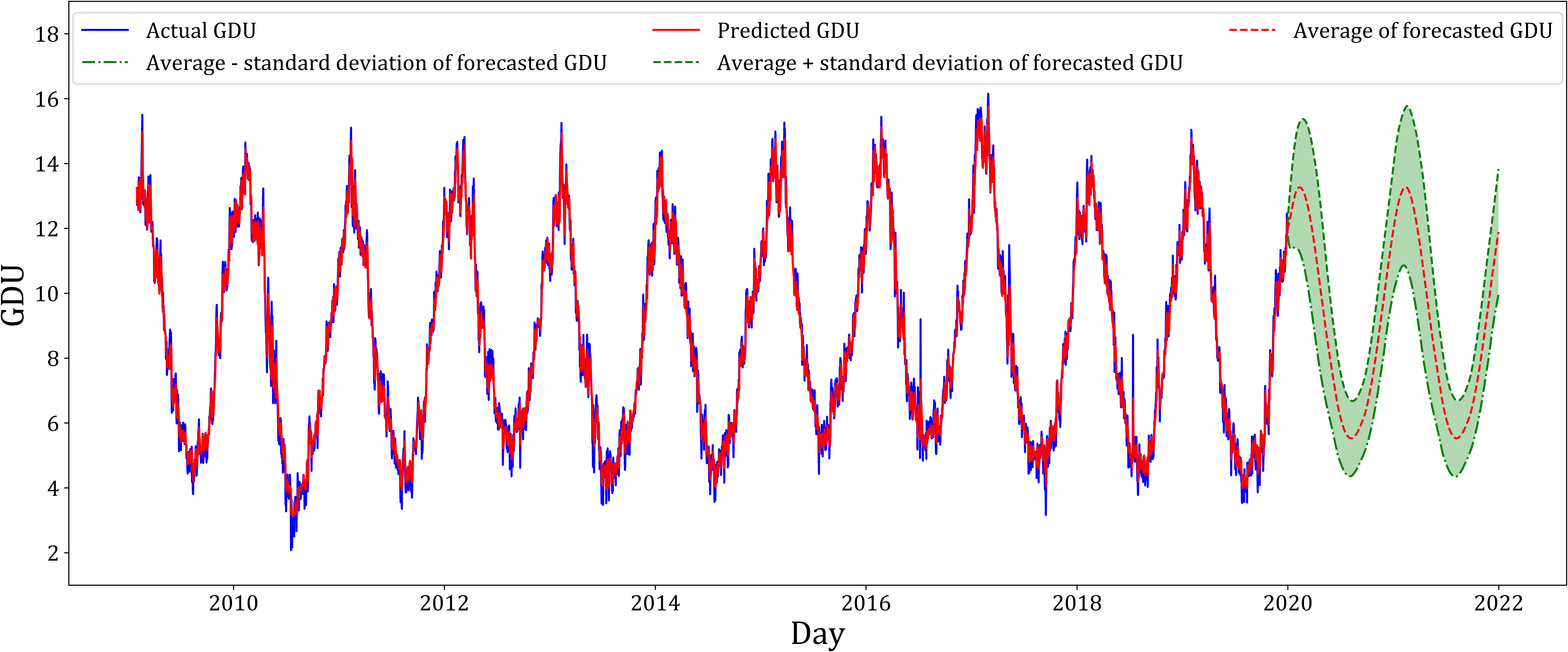}
	\caption{Forecasted GDU and its uncertainty at site 0 into the future (years 2020 and 2021) using RIO algorithm.}\label{LSTMSite0p}
\end{figure}

\begin{figure}[H]
	\includegraphics[height=2in]{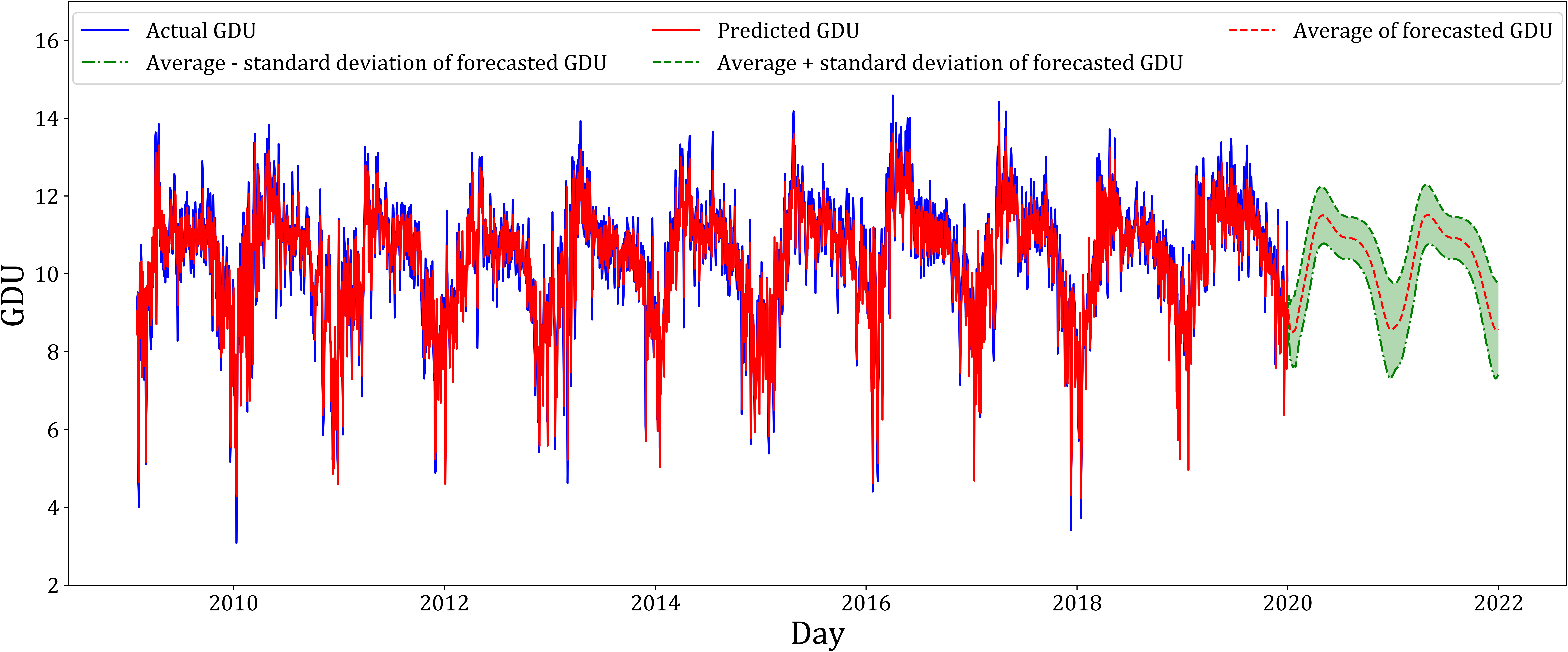}
	\caption{Forecasted GDU and its uncertainty at site 1 into the future (years 2020 and 2021) using RIO algorithm.}\label{LSTMSite1p}
\end{figure}

The results of using the RIO to estimate uncertainty in the prediction of the GDU into the future (next two years) for sites 0 and 1 were visualized in Figure \ref{LSTMSite0p} and \ref{LSTMSite1p}, respectively. The proposed time-series model was trained on data up to the end of 2019, and the predictions started on the first of 2020, and it then predicted the GDU 730 days into the future. To model uncertainty of weather for the next planting and harvesting calendar (2020 and 2021), the RIO approach was used to create 25 weather scenarios. The shadow areas represent the confidence interval of weather prediction, which indicates the range of variability across 25 weather scenarios. We used 25 weather scenarios to formulate the stochastic optimization model under weather uncertainty on the given calendar day of 2020.

The predicted GDU can be compared to the actual GDU during the historical period (2009-2019), and thus the forecasted GDUs into the future follow meaningful trajectories. This result can be attributed to sampling from the Gaussian distribution via Monte Carlo rollouts to estimate weather into the future instead of predicting only by the proposed time-series model and feeding it back into the model to predict the next step. 

\subsection{Optimal schedule for planting time of seed population}

\begin{figure}[H]
	\centering
	\begin{picture}(50,150)(80,0)
		\put(120,0){\includegraphics[width=2.25in]{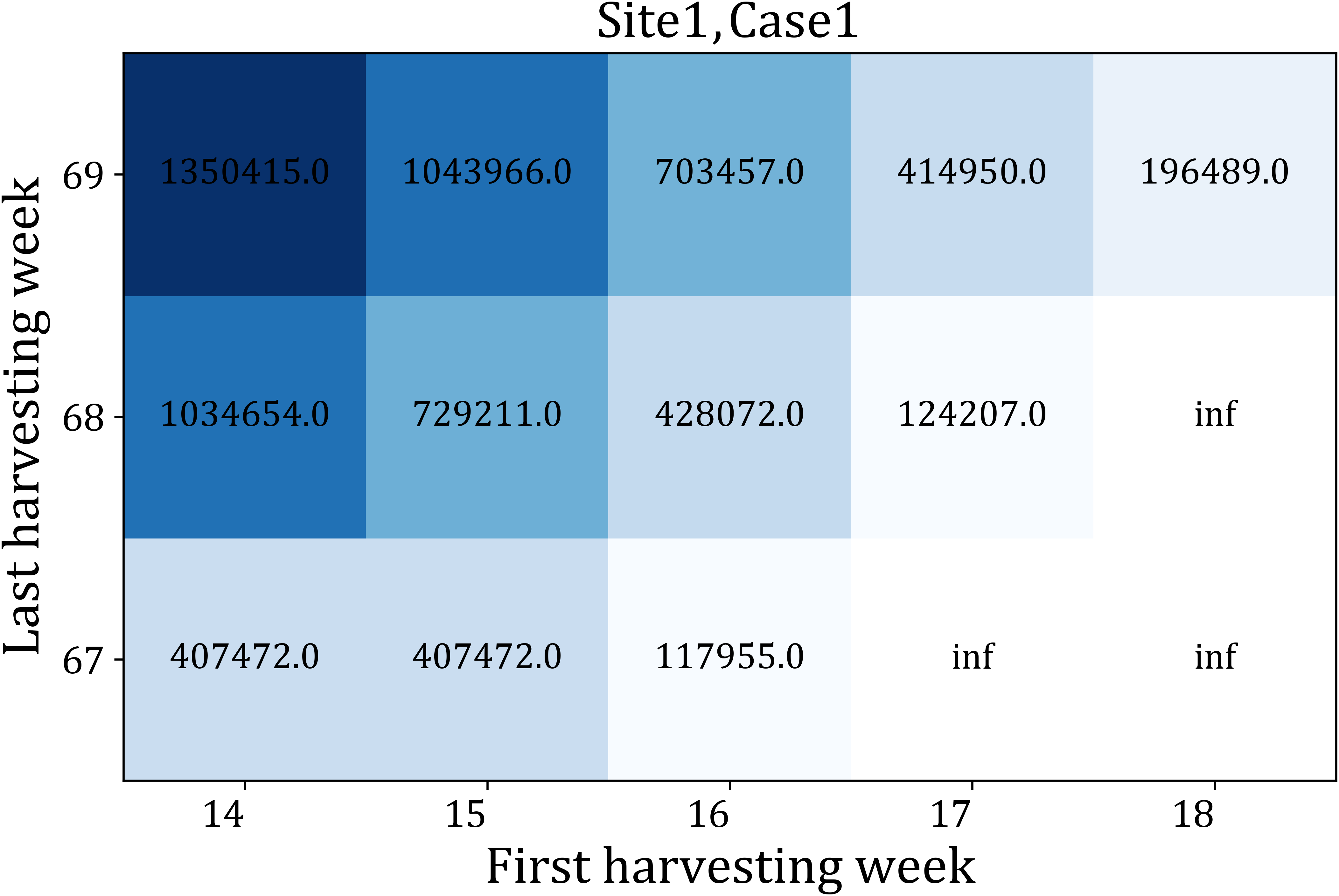}}
		\put(-125,0){\includegraphics[width=2.25in]{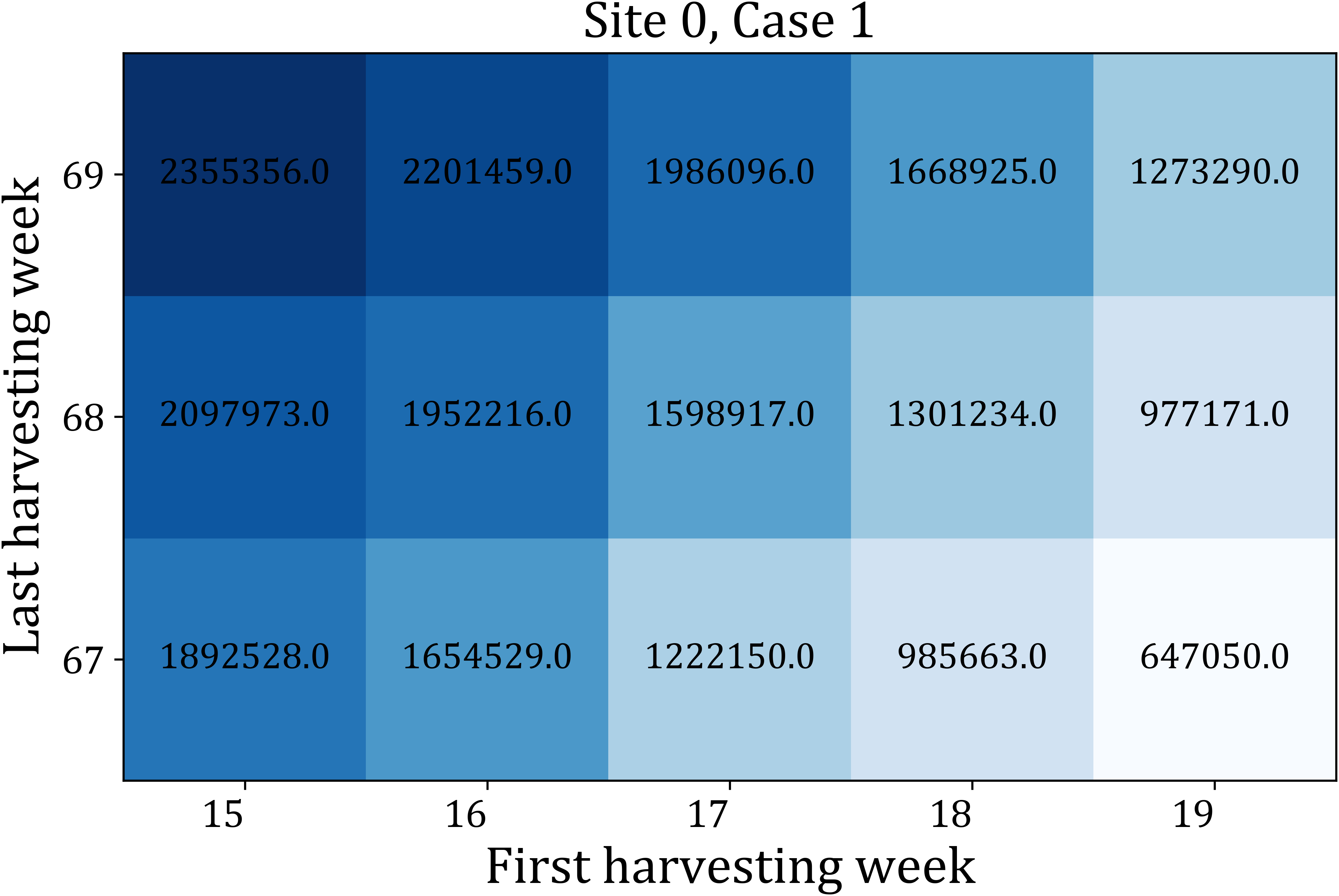}}
	\end{picture}
	\caption{The objective values of Equation (\ref{new}) for the different allowed harvesting periods at sites 0 and 1. The inf value refers to the infeasibility of the optimization model (\ref{obj1})-(\ref{con3}). The numbers in each block refer to the value of Equation (\ref{new}). The darker blocks have higher objective function values, and they are not optimal.}\label{selection1}
\end{figure}

\begin{figure}[H]
	\includegraphics[height=2.5in]{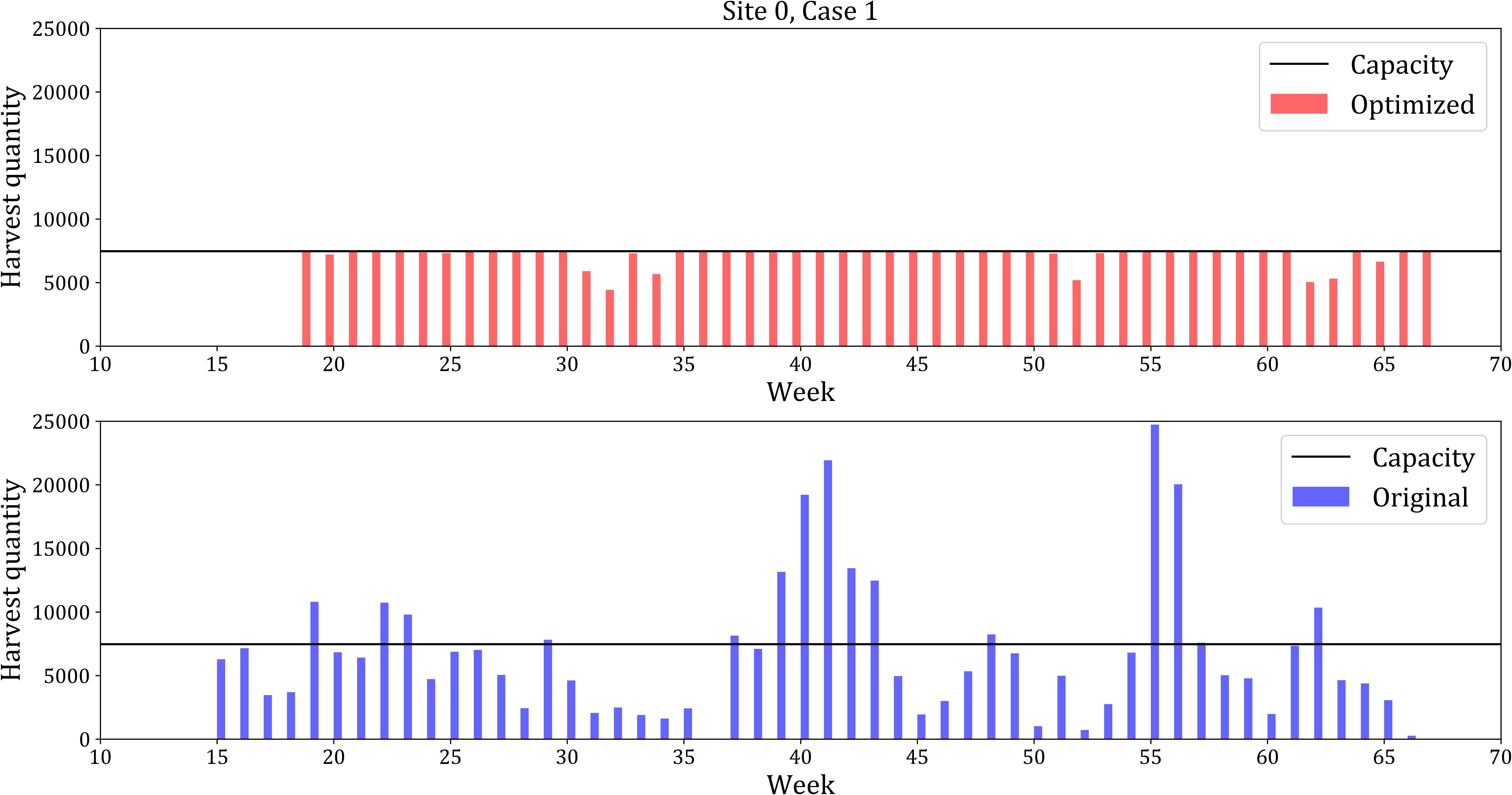}
	\caption{The original and optimal weekly harvest quantities at site 0 in case 1 using the average of forecasted GDU.}\label{Site0_Case1}
\end{figure}

\begin{figure}[H]
	\includegraphics[height=2.5in]{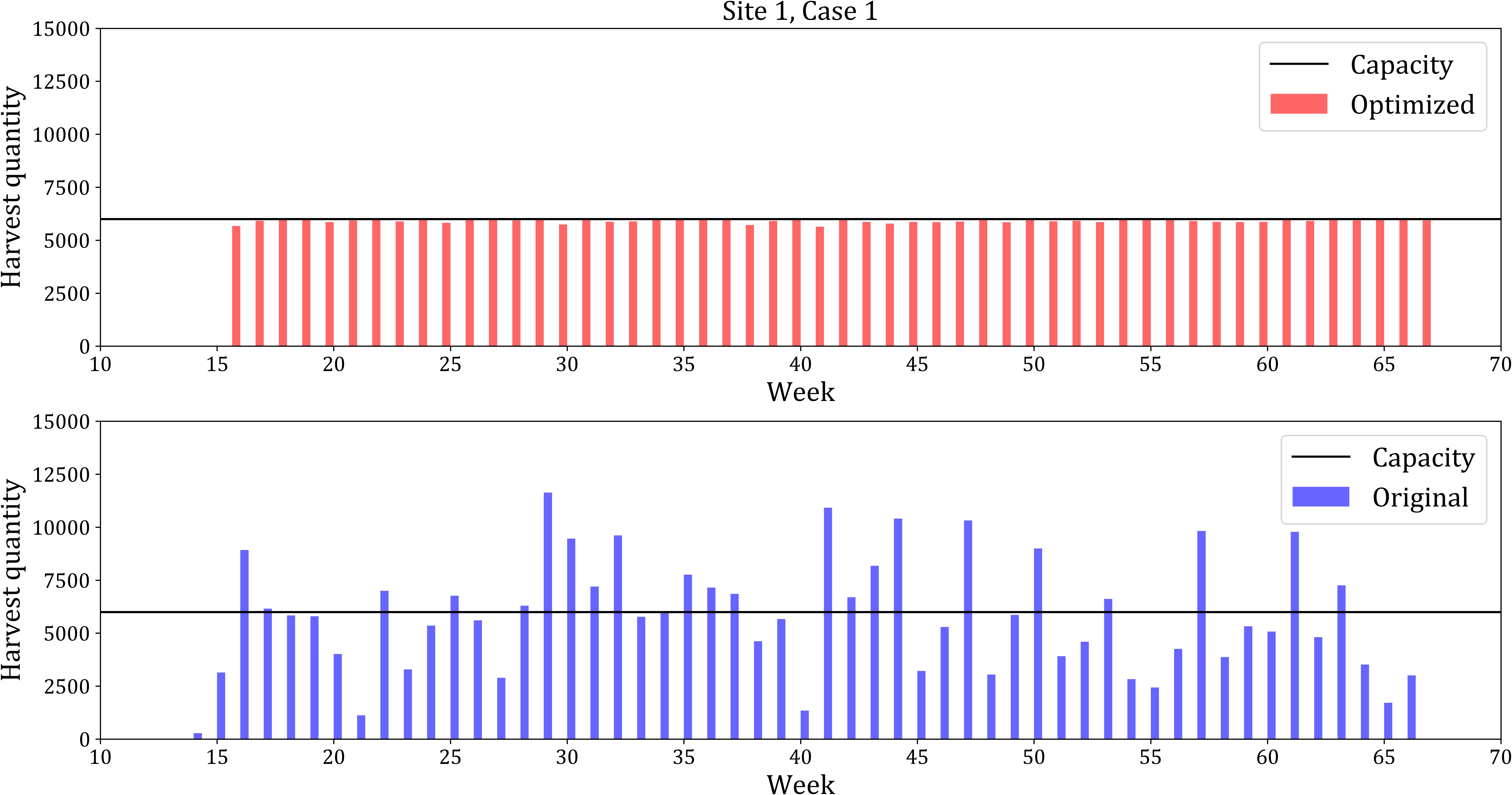}
	\caption{The original and optimal weekly harvest quantities at site 1 in case 1 using the average of forecasted GDU.}\label{Site1_Case1}
\end{figure}

The optimization model's goal in case 1 (sites have capacities) is to schedule the seed population's planting time within the fewest number of weeks. Hence, we iteratively limited the first and the last week harvest weeks to the specific shorter periods for two sites, and then the optimization model (\ref{obj1})-(\ref{con3}) scheduled the best planting time for seed populations so that the harvest must be done in these predefined periods. Figure \ref{selection1} shows the objective values of Equation (\ref{new}) for the various harvesting periods at sites 0 and 1. The results show that the best planting dates with the highest consistent weekly harvest quantity and the fewest harvest weeks are when the allowed harvesting weeks are week 19 to week 67 for site 0 and from week 16 to week 67 for site 1. Figures \ref{Site0_Case1} and \ref{Site1_Case1} illustrate the original and optimal weekly harvest quantities at sites 0 and 1 in case 1 using the average of forecasted GDU. Table \ref{Stat2} reports the maximum required capacity, harvesting period, and value of Equation (\ref{new}). Our proposed model decreased the required capacity by 69\% at site 0 and 48\% at site 1 compared to the original planting time. Also, the proposed approach reduced the harvesting period by 1 week. 

\begin{table}[H]
	\centering
	\caption{Optimal and original planting times for case 1.}\label{Stat2}
	\scalebox{0.8}{
		\begin{tabular}{cc|c|c|c|c}
			\clineB{1-6}{4.5}
			\multirow{4}{*}{Method}&\multicolumn{5}{c}{Site 0}\\
			\hhline{~-----}
			&{\makecell{Objective \\function}} &{\makecell{First\\ harvesting\\ time}}  &{\makecell{Last\\ harvesting\\ time}}  &{\makecell{Harvesting\\ period}}  &{\makecell{Maximum \\ required\\capacity}}\\
			\hline
			Original&7,410,283&15&66&52&24,736\\
			Optimal&647,050&19&67&49&7,475\\
			\clineB{1-6}{4.5}
			\multirow{4}{*}{Method}&\multicolumn{5}{c}{Site 1}\\
			\hhline{~-----}
			&{\makecell{Objective \\function}} &{\makecell{First\\ harvesting\\ time}}  &{\makecell{Last\\ harvesting\\ time}}  &{\makecell{Harvesting\\ period}}  &{\makecell{Maximum \\ required\\capacity}}\\
			\hline
			Original&4,263,080&14&66&53&11,632\\
			Optimal&117,955&16&67&52&6,000\\
			\clineB{1-6}{4.5}
	\end{tabular}}
\end{table}

\begin{figure}[H]
	\centering
	\begin{picture}(50,120)(80,0)
		\put(120,0){\includegraphics[width=1.75in]{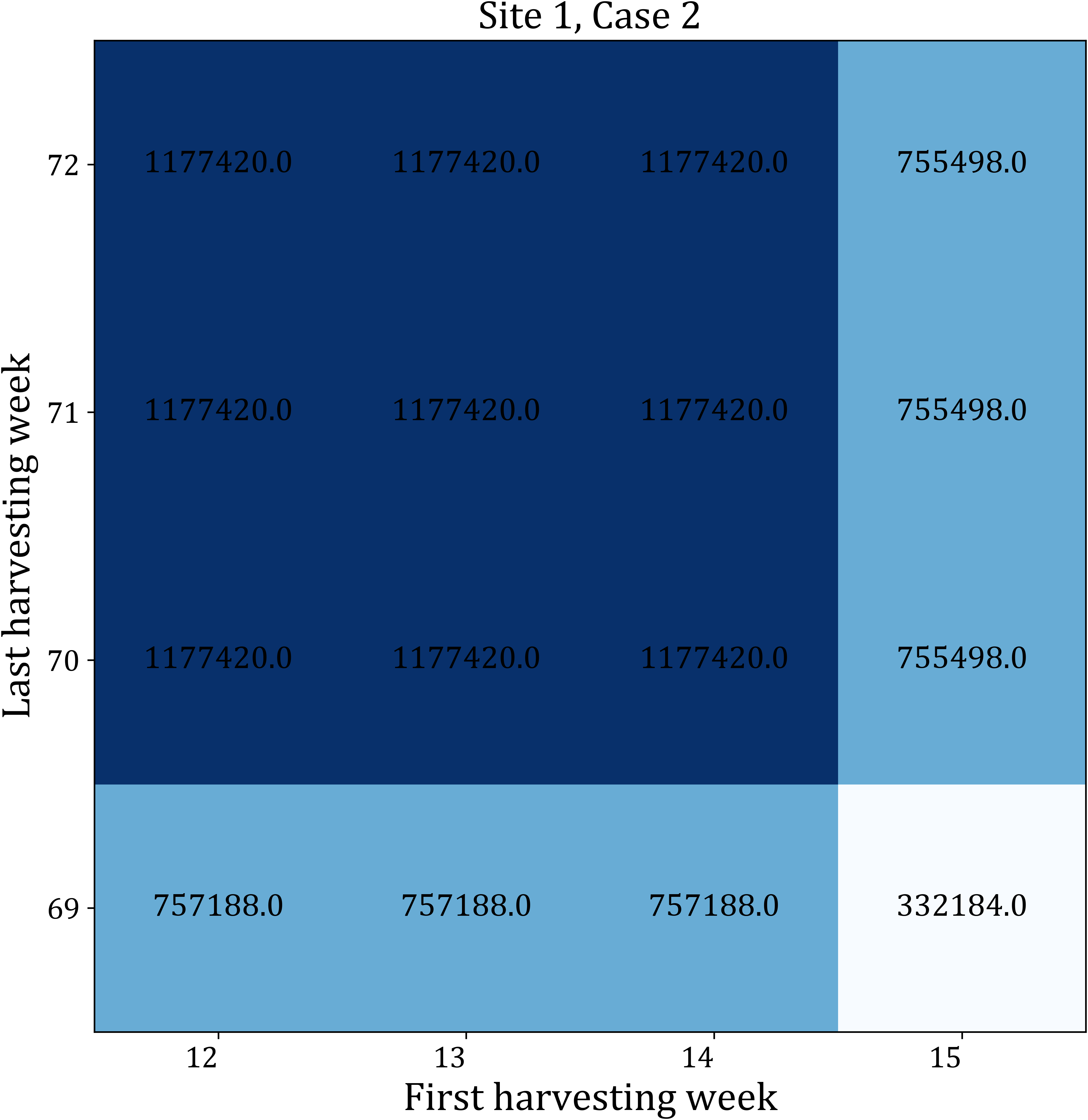}}
		\put(-125,0){\includegraphics[width=2in]{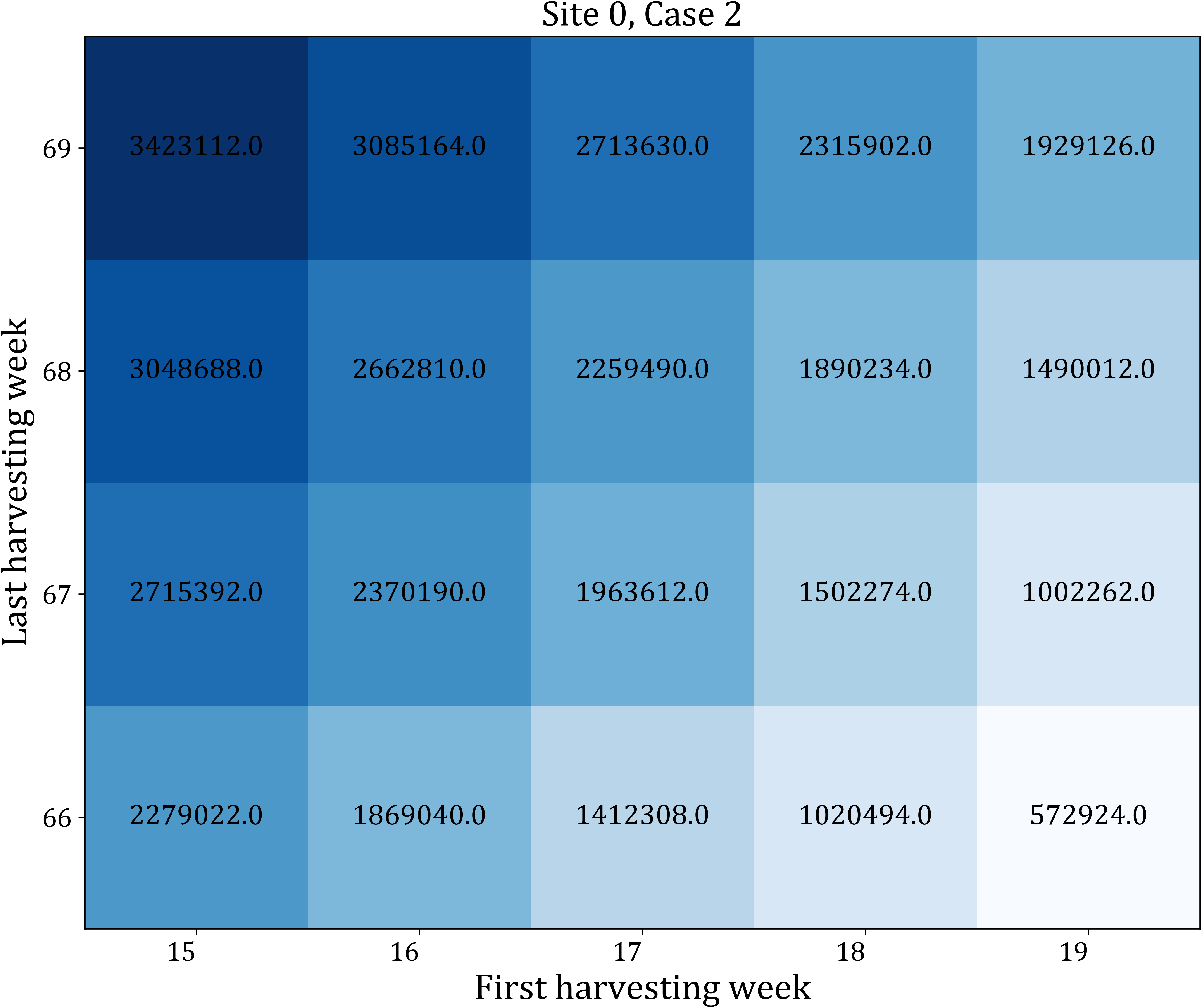}}
	\end{picture}
	\caption{The objective values of Equation (\ref{new}) for the different allowed harvesting periods at sites 0 and 1. The numbers in each block refers to value of Equation (\ref{new}). The darker blocks have higher objective function values, and they are not optimal.}\label{selection2}
\end{figure}

The aim of the optimization model in case 2 (sites have no capacities) is to schedule the seed population's planting time at the lowest capacity required for both sites as well as the fewest number of weeks. Solving the optimization model (\ref{obj2})-(\ref{con6}) for both sites suggested that the lowest capacity required for sites 0 and 1 are 10,658 and 7,875. Then, the optimization model (\ref{obj1})-(\ref{con3}) was solved for various harvesting periods with determined capacities. For this purpose, we limited the model to determine the seeds' planting times so that their harvests happened in the limited harvesting periods. The best harvesting periods in terms of minimizing Equation (\ref{new}) are reported in Figure \ref{selection2} for sites 0 and 1.

\begin{figure}[H]
	\includegraphics[height=2.5in]{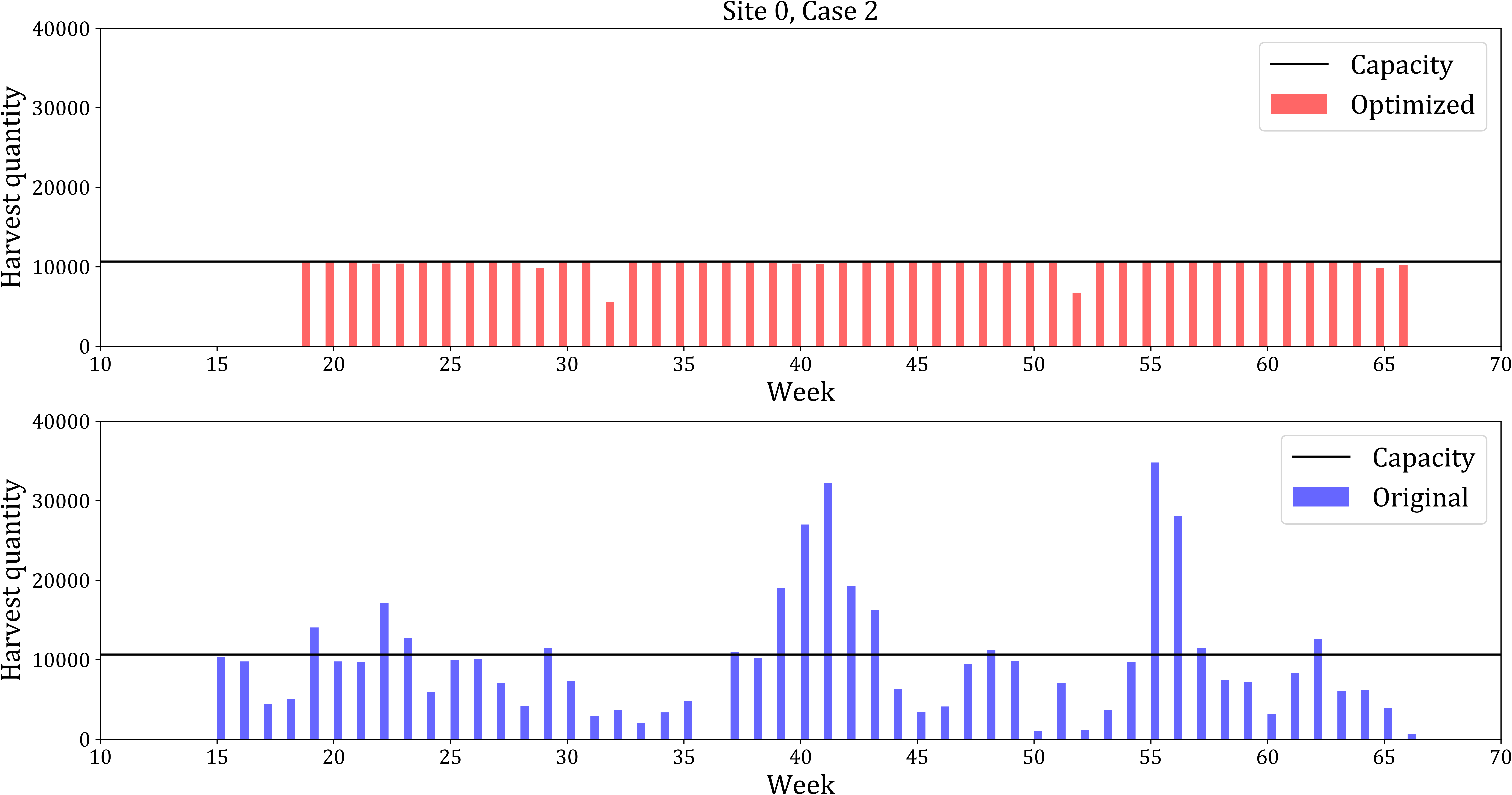}
	\caption{The original and optimal weekly harvest quantities at site 0 in case 2 using the average of forecasted GDU.}\label{Site0_Case2}
\end{figure}

\begin{figure}[H]
	\includegraphics[height=2.5in]{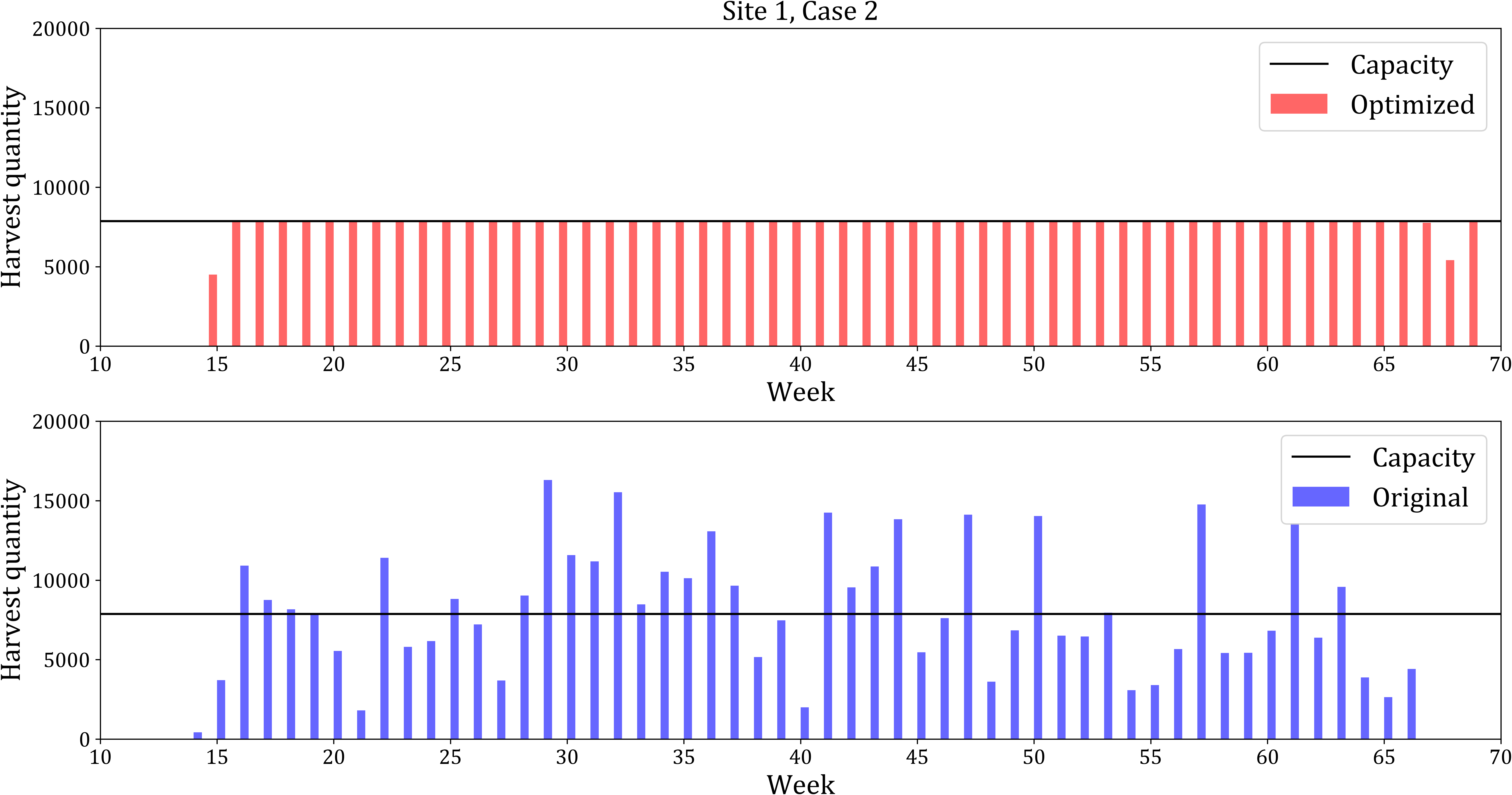}
	\caption{The original and optimal weekly harvest quantities at site 1 in case 2 using the average of forecasted GDU.}\label{Site1_Case2}
\end{figure}

The best harvesting period for site 0 is week 19 to week 66 and for site 1 is week 15 to week 69. The results of solving the optimization model (\ref{obj1})-(\ref{con3}) with determined capacities from model (\ref{obj2})-(\ref{con6}) and optimal harvest week for both sites are shown in Figures \ref{Site0_Case2} and \ref{Site1_Case2}. These figures indicate the original and optimal weekly harvest quantities at sites 0 and 1 in case 2 using the average of forecasted GDU. Table \ref{Stat3} shows that our proposed model found the lowest required capacities by decreasing the capacity by 69\% at site 0 and by 51\% at site 1.

\begin{table}[H]
	\centering
	\caption{Optimal and original planting times for case 2.}\label{Stat3}
	\scalebox{0.8}{
		\begin{tabular}{cc|c|c|c|c}
			\clineB{1-6}{4.5}
			\multirow{4}{*}{Method}&\multicolumn{5}{c}{Site 0}\\
			\hhline{~-----}
			&{\makecell{Objective \\function}} &{\makecell{First\\ harvesting\\ time}}  &{\makecell{Last\\ harvesting\\ time}}  &{\makecell{Harvesting\\ period}}  &{\makecell{Maximum \\ required\\capacity}}\\
			\hline
			Original&10,362,758&15&66&52&34,799\\
			Optimal&572,924&19&66&48&10,658\\
			\clineB{1-6}{4.5}
			\multirow{4}{*}{Method}&\multicolumn{5}{c}{Site 1}\\
			\hhline{~-----}
			&{\makecell{Objective \\function}} &{\makecell{First\\ harvesting\\ time}}  &{\makecell{Last\\ harvesting\\ time}}  &{\makecell{Harvesting\\ period}}  &{\makecell{Maximum \\ required\\capacity}}\\
			\hline
			Original&6,210,306&14&66&53&16,299\\
			Optimal&332,184&15&69&55&7,875\\
			\clineB{1-6}{4.5}
	\end{tabular}}
\end{table}

\section{Conclusion}

We developed a new framework with the combination of time-series model and optimization models to address the 2021 Syngenta crop challenge by scheduling the planting time of seed populations at the lowest capacity required and the fewest number of harvest weeks. This challenge is to optimize the seed populations' planting times during 2020. Hence, the unseen weather information at the given calendar days was forecasted by the proposed time-series model that consists of the LSTM model and fully connected deep learning. To estimate the uncertainty of the weather forecast into the future, we used the RIO model. The results reported the forecasted weather follows historical trajectories. By having the weather scenarios, we proposed a stochastic optimization model to schedule the farming system. Results from the computational experiment suggested that the optimization model achieved a more consistent weekly harvest quantity in fewer harvesting weeks.

\section{Data availability}

The data analyzed in this study was provided by Syngenta for the 2021 Syngenta crop challenge. We accessed the data through the annual Syngenta crop challenge. During the challenge, September 2020 to January 2021, the data was open to the public. Researchers who wish to access the data may do so by contacting Syngenta directly (https://www.ideaconnection.com/syngenta-crop-challenge/challenge.php).

\section{Conflict of Interest Statement}

The authors declare that the research was conducted in the absence of any commercial or financial relationships that could be construed as a potential conflict of interest.

\section{Funding}

This work was supported by the National Science Foundation under the LEAP HI and GOALI programs (grant number 1830478) and under the EAGER program (grant number 1842097).

\section{Acknowledgments}

The authors are thankful to Syngenta and the Analytics Society of INFORMS for organizing the 2021 Syngenta crop challenge and sharing the invaluable dataset with the research community. 

\section{Author Contributions}

J.A. and F.A. conducted the research by preparing and cleaning the database, designing and implementing algorithms, performing the experiment and statistical analysis, interpreting experiment results, and writing the manuscript. L.W. oversaw the research, reviewed, and edited the manuscript. All authors read and approved the final manuscript.

\bibliographystyle{unsrt}  
\bibliography{references}

\end{document}